\documentclass[11pt]{article}

\usepackage[final]{acl}

\usepackage{times}
\usepackage{latexsym}

\usepackage[T1]{fontenc}

\usepackage[utf8]{inputenc}

\usepackage{microtype}

\usepackage{inconsolata}

\usepackage{graphicx}
\usepackage{amsmath}
\usepackage{float}
\usepackage{mathrsfs}
\usepackage{amssymb} 
\usepackage[linesnumbered,ruled,vlined]{algorithm2e}
\usepackage[most]{tcolorbox}
\usepackage{xcolor}
\usepackage{listings}
\usepackage{tcolorbox}
\usepackage{enumitem}
\usepackage{dsfont}
\usepackage{multirow}
\usepackage{makecell}
\usepackage{threeparttable}  
\usepackage{arydshln} 
\usepackage{float}  
\usepackage{caption}
\usepackage{booktabs}
\usepackage[table]{xcolor}
\usepackage{booktabs, arydshln}
\usepackage{tikz}
\usetikzlibrary{shapes,arrows,positioning}
\newtheorem{proposition}{Proposition}

\title{Every Step Counts: Step-Level Credit Assignment for Tool-Integrated Text-to-SQL}

  \author{
  Yaxun~Dai\textsuperscript{\rm 1,2},
  ~Baolin Sun\textsuperscript{\rm 2}\footnotemark[1],
  ~Junying Wang\textsuperscript{\rm 3},
  ~Pengfei Wang\textsuperscript{\rm 2},
  ~Yingqi Gao\textsuperscript{\rm 2}, \\
  ~\textbf{Xuemei Dong}\textsuperscript{\rm 2},
  ~\textbf{Mengdie Chu}\textsuperscript{\rm 2},
  ~\textbf{Xiang Qi}\textsuperscript{\rm 2}\footnotemark[1],
  ~\textbf{Pingfu~Chao}\textsuperscript{\rm 1}\footnotemark[1] \\[6pt]
\textsuperscript{\rm 1}Institute of Computer Science and Technology, Soochow University, China, \\
\textsuperscript{\rm 2}Ant Digital Technologies, Ant Group, \\
\textsuperscript{\rm 3}School of Management, University of Science and Technology of China\\
\{daiyaxun.dyx,xuanfeng.sbl,qixiang.qx\}@antgroup.com,~~pfchao@suda.edu.cn
}

\begin{document}
\maketitle

{
  \renewcommand{\thefootnote}{\fnsymbol{footnote}}
  \footnotetext[1]{Corresponding authors.}
}

\begin{abstract}
Tool-integrated Text-to-SQL parsing has emerged as a promising paradigm, framing SQL generation as a sequential decision-making process interleaved with tool execution. 
However, existing reinforcement learning approaches mainly rely on coarse-grained outcome supervision, resulting in a fundamental credit assignment problem: models receive the same reward for any trajectory that yields the correct answer, even when intermediate steps are redundant, inefficient, or erroneous. 
Consequently, models are encouraged to explore suboptimal reasoning spaces, limiting both efficiency and generalization.
To address this problem, we propose \textbf{FineStep}, a novel framework for step-level credit assignment in tool-augmented Text-to-SQL. 
First, we introduce a reward design with independent process rewards to alleviate the signal sparsity of outcome supervision. 
Next, we present a step-level credit assignment mechanism to precisely quantify the value of each reasoning step. 
Finally, we develop a policy optimization method based on step-level advantages for efficient updates.
Extensive experiments on BIRD benchmarks show that FineStep achieves state-of-the-art performance and reduces redundant tool interactions, with a 3.25\% average EX gain over GRPO at the 4B scale.
\end{abstract}

\section{Introduction}
\label{sec:intro}


The Text-to-SQL task, which aims to translate natural language questions into executable SQL queries, serves as a critical bridge connecting non-expert users with structured data~\citep{DBLP:journals/corr/surveynl2sql,DBLP:journals/TKDE/ZhangWSWTQCWY24}.
With the rapid advancement of Large Language Models (LLMs), the research paradigm in this field is undergoing a profound shift from single-step semantic parsing to {Tool-augmented Agentic Reasoning~\citep{singh2025agenticreasoningtoolintegration,mai2025agentrlscalinglaw,Agentar—SQL}}.
Under this paradigm, the model no longer generates the final query in a single pass but iteratively approaches the solution through multi-turn interactions with an {SQL executor}: generating SQLs, observing execution feedback, and refining logic in a loop until the correct answer is derived~\citep{reex-sql}.


\begin{figure}[t]
    \includegraphics[width=1\columnwidth]{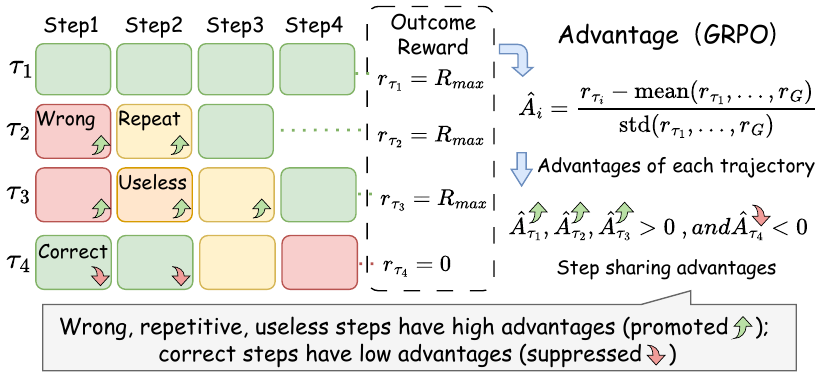}
    \caption{The credit assignment problem in standard GRPO~\citep{GRPO}. $\tau$ denotes a reasoning trajectory whose final step outputs the answer. 
Colors mark \textcolor{red!60}{wrong}, \textcolor{yellow!90!black}{repetitive}, \textcolor{orange}{useless}, and \textcolor{green!60!black}{correct} steps, identified via tool calls in tool-integrated reasoning.}
    \label{fig:teaser}
\end{figure}

To optimize decision-making strategies within this complex search space, Agentic Reinforcement Learning (Agentic RL)~\citep{agenticrl-survey} has become the dominant paradigm for enhancing multi-step reasoning capabilities~\citep{rStar2-Agent,arpo}.
{However}, existing methods (e.g., GRPO~\citep{GRPO}) update policies relying solely on rewards derived from the final execution result~\citep{reex-sql,mtir-sql}. This mechanism faces a severe {Credit Assignment Problem} in long-horizon reasoning tasks, which refers to the difficulty of quantifying the contribution of intermediate steps~\citep{gigpo,toolrl-scaling,simpletir}.
Due to sparse feedback, this mechanism rewards the trajectory as a whole, {consequently failing to distinguish between effective reasoning steps and irrelevant redundant operations}~\citep{ji2025treesearchllmagent,lai2025computerrlscalingendtoendonline,agentrm}.
As illustrated in Figure~\ref{fig:teaser}, a trajectory filled with redundant or even erroneous steps receives high rewards solely because the final answer is correct.
This misleading signal induces the model to learn inefficient ``blind trial-and-error'' strategies rather than genuine logical reasoning, thereby severely compromising robustness~\citep{step2025successrateaware,agentrm}.

To break the ``trajectory-level'' black box and explicitly quantify intermediate step utility, we propose \textbf{FineStep}, a fine-grained credit assignment framework tailored for tool-augmented Text-to-SQL. 
First, alongside outcome supervision, we introduce {independent process rewards} to evaluate step validity and information gain.
Second, we design a {multi-dimensional step-level credit assignment strategy} to transform raw rewards into discriminative advantages across three dimensions:
(1) employing {outcome reward discounting} to address temporal heterogeneity and reinforce the causal links of recent actions;
(2) {process reward smoothing} to capture logical dependencies and affirm the instrumental value of ``preparatory steps'' with low immediate returns;
(3) {group-based step-level normalization} to eliminate baseline discrepancies, identifying ``correct steps in failed trajectories'' and ``redundancies in successful ones.''
Finally, we extend GRPO to the {step level} and incorporate a {masking mechanism} for external feedback, ensuring the model focuses strictly on optimizing its internal reasoning logic.
In summary, the main contributions of this paper are as follows:


\begin{itemize}[leftmargin=1em]
    \item We propose \textbf{FineStep}, a framework that augments coarse-grained outcome supervision with independent process rewards to explicitly evaluate intermediate reasoning utility, overcoming the limitations of sparse feedback.
    \item We design a multi-dimensional {step-level credit assignment mechanism} that addresses the attribution challenge in long-horizon reasoning by reconstructing advantage estimation across temporal, logical, and group-relative dimensions.
    \item Extensive experiments across benchmarks (e.g., BIRD) show that FineStep surpasses SOTA execution accuracy while reducing reasoning steps and tool invocation errors, achieving a 3.25\% average EX gain over GRPO at the 4B scale.
\end{itemize}

\section{Related Work}


\subsection{LLM Reasoning in Text-to-SQL}
With the advancement of LLMs, the research focus in Text-to-SQL has shifted from traditional semantic parsing to complex reasoning generation~\citep{DBLP:journals/corr/surveynl2sql,DBLP:journals/corr/liu2024survey}.
Early efforts primarily utilized Prompt Engineering (e.g., CoT~\citep{DBLP:conf/nips/cot}, DIN-SQL~\citep{DBLP:conf/nips/din_sql}, Alpha-SQL~\citep{Alpha-SQL}, CHASE-SQL~\citep{pourreza2024chase}) to guide reasoning via pre-orchestrated static workflows. 
However, these methods lack the capability to autonomously optimize strategies.
To address this, Reinforcement Learning was introduced. 
Representative works such as {Reasoning-SQL}~\citep{reasoning-sql}, {SQL-R1}~\citep{sql-r1}, and {Arctic-Text2SQL-R1}~\citep{Arctic-sql} treat the model as a single-step decision-maker, fine-tuning it using final execution accuracy.
However, this single-turn paradigm lacks self-correction capabilities, limiting its effectiveness on complex queries~\citep{reex-sql,mars-sql}.
Consequently, the field has shifted towards {Tool-integrated Agentic Reasoning}.
Frameworks like {ReEx-SQL~\citep{reex-sql}}, {MTIR-SQL~\citep{mtir-sql}}, {SKYRL-SQL~\citep{liu2025skyrlsql}}, and {MARS-SQL~\citep{mars-sql}} formulate SQL generation as an interactive process, enabling iterative refinement through a ``Model-Tool-Feedback'' loop.
{Despite this shift, existing optimization methods still predominantly rely on coarse-grained outcome supervision~\citep{GRPO,dapo}}: as long as the final SQL is correct, the entire trajectory—including potential redundant or erroneous steps—receives positive reinforcement.
This failure to distinguish intermediate step quality leads to the erroneous reinforcement of invalid parsing or redundant invocations, highlighting a critical limitation in {credit assignment} for multi-turn Text-to-SQL.


\subsection{Credit Assignment in LLM Reasoning}
To mitigate feedback sparsity, Process Supervision has garnered widespread attention~\citep{Rewarding-Progress,reward-sql}.
For instance, research in mathematical reasoning~\citep{Math-Shepherd,PRM800K,deepseekr1} has validated the superiority of step-based dense rewards, while works in the web agent domain~\citep{gigpo,step2025successrateaware,fan2026exploringreasoningrewardmodel} have explored step-level credit assignment via time-decay mechanisms.
{However}, integrating process supervision into Agentic RL remains a nascent direction~\citep{agenticrl-survey}.
Existing methods typically rely on expensive human annotation (e.g., PRMs~\citep{lessons-prms}) or employ simple heuristic decay~\citep{gigpo}, failing to capture the unique logical dependencies in Text-to-SQL, such as the instrumental value of \textit{exploratory queries} for subsequent generation~\citep{DBLP:journals/TKDE/ZhangWSWTQCWY24}.
Notably, existing PRM-based approaches such as PRIME~\citep{PRIME} and Reward-SQL~\citep{reward-sql} are not directly aligned with our setting. Reward-SQL focuses on internal step-wise Text-to-SQL reasoning, whereas FineStep targets \textit{tool-integrated multi-turn Text-to-SQL}, where each step comprises reasoning, an SQL tool call, and executor feedback. Consequently, existing PRMs cannot be directly transferred to tool-integrated reasoning (TIR) setting. Constructing a PRM-based TIR baseline would require building tool-interaction-level PRM data, performing additional SFT/PRM training, and deploying the PRM during RL or inference—introducing extra supervision, computation, and experimental variables.

\section{Methodology}
\label{sec:method}

\begin{figure}[t]
\footnotesize
\begin{tcolorbox}[colback=yellow!1.5, boxrule=0.5pt]
\textbf{Response Format:} \\
\textcolor[HTML]{2e57d0}{{<reasoning>}} Reasoning process here. \textcolor[HTML]{2e57d0}{{</reasoning>}} \\
\textcolor[HTML]{ffb713}{{<tool\_call>}} \{"name": "sql\_executor", "arguments": \{"sql": "YOUR\_SQL"\}\} \textcolor[HTML]{ffb713}{{</tool\_call>}} \\
\textcolor[HTML]{16b69e}{{<result>}} Execution result from the engine. \textcolor[HTML]{16b69e}{{</result>}} \\
\textcolor[HTML]{2e57d0}{{<reasoning>}} Further reasoning. \textcolor[HTML]{2e57d0}{{</reasoning>}} \\
\textcolor[HTML]{bf0040}{{<answer>}} (Final verified SQL) \textcolor[HTML]{bf0040}{{</answer>}}
\end{tcolorbox}
\caption{Response format for FineStep.}
\label{prompt:reasoning_format}
\end{figure}

\begin{figure*}[t]
    \includegraphics[width=1\textwidth]{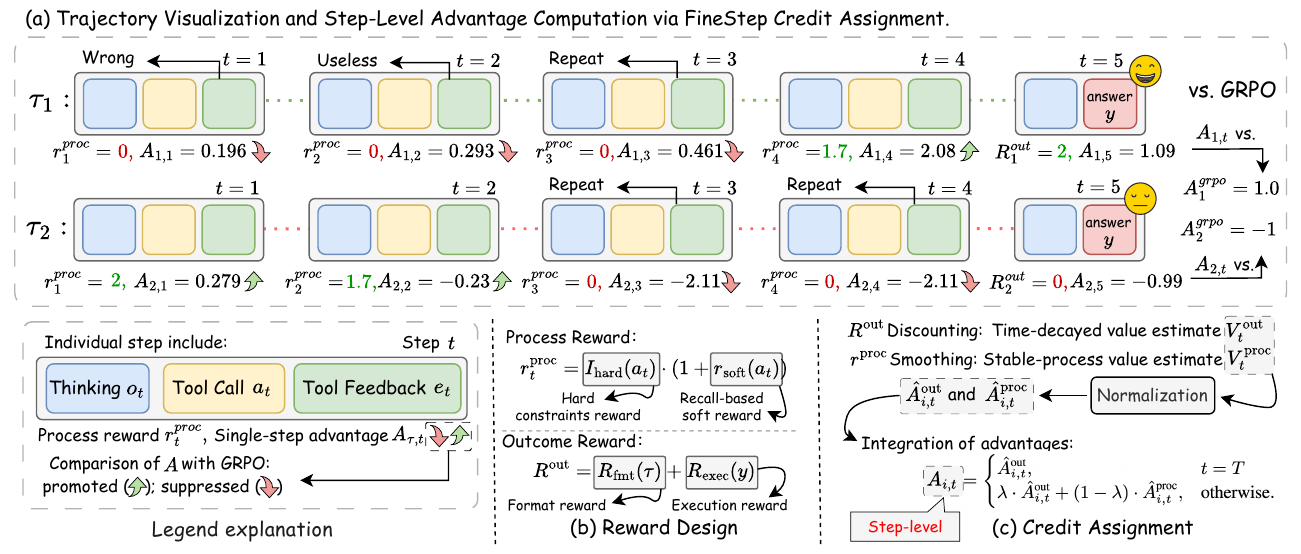}
    \caption{Overview of FineStep, illustrating its step-level credit assignment and reward design. (a) Trajectory visualization and step-level advantage computation, showing how FineStep promotes or suppresses specific reasoning steps compared to GRPO; (b) reward design with process and outcome rewards; and (c) credit assignment integrating discounted outcome and smoothed process advantages. More details are provided in Section~\ref{sec:method}.}
    \label{fig:finestep}
\end{figure*}



To address the {Credit Assignment Problem} in tool-augmented Text-to-SQL, we propose \textbf{FineStep}, a novel framework for step-level credit assignment in Figure~\ref{fig:finestep}.
{First}, we introduce the {Reward Design}, incorporating independent process rewards to mitigate signal sparsity caused by outcome supervision.
{Next}, we detail the core {Step-level Credit Assignment Mechanism}, aimed at precisely quantifying the true value of each reasoning step.
{Finally}, we present the {policy optimization} based on step-level advantages to achieve efficient parameter updates.


\subsection{Problem Definition}
We formulate the tool-augmented Text-to-SQL task as a sequential decision-making process interacting with an SQL executor.
Given a natural language question $x$ and a database schema $\mathcal{S}$, the initial state is denoted as $s_{\text{init}} = \{x, \mathcal{S}\}$.
As illustrated in Figure~\ref{prompt:reasoning_format}, the model adheres to a ``Think-Act-Observe'' paradigm.
Here, we explicitly define a single \textbf{reasoning step $t$} as a complete interaction cycle:
the model's thought $o_t$, action $a_t$, and the execution feedback $e_t$.
Finally, the model autonomously generates the final answer $y$.
The resulting trajectory $\tau$ of length $T+1$ is:
\begin{equation}
    \tau = \{s_{\text{init}}, o_0, a_0, e_0, \dots, o_{T-1}, a_{T-1}, e_{T-1}, o_{T}, y\}
\end{equation}


\subsection{Reward Design}
To achieve precise credit assignment, we construct a multi-granular reward function comprising an outcome reward ($R^{\text{out}}$) and a process reward ($r_t^{\text{proc}}$).
The former evaluates the final correctness to provide global guidance, while the latter quantifies step-wise information gain, effectively mitigating the signal sparsity in outcome supervision.



\subsubsection{Outcome Reward Design}
The outcome reward evaluates the global consistency of the reasoning trajectory and the correctness of the final answer, defined as the sum of format and execution rewards:
\begin{equation}
    R^{\text{out}} = R_{\text{fmt}}(\tau) + R_{\text{exec}}(y)
\end{equation}

Specifically, we define $R_{\text{fmt}} \in \{0, 1\}$ to enforce the structural paradigm in Figure~\ref{prompt:reasoning_format} (e.g., \texttt{<think>}  and \texttt{<answer>} tokens), and $R_{\text{exec}} \in \{0, 2\}$ to evaluates whether the execution result of the final SQL $y$ exactly matches the ground truth, adopting the higher weighting strategy from ReEx-SQL~\citep{reex-sql}.



\subsubsection{Process Reward Design}

To mitigate the signal sparsity inherent in outcome supervision, we formulate the fine-grained process reward $r_t^{\text{proc}}$ as a combination of {hard constraints on validity} and {soft rewards for information gain}.


\paragraph{Hard Constraints Reward}
Invalid steps (e.g., erroneous tool calls) not only provide {zero information gain} but also {impair the reasoning process} by introducing noise and unnecessarily {lengthening the chain of thought}.
To mitigate this, we design a rule-based hard constraint $I_{\text{hard}}$ to assign the minimal reward to non-compliant steps:
\begin{equation}
    I_{\text{hard}}(a_t) = 
    \begin{cases} 
    0, & \text{if } a_t \text{ is invalid or repetitive} \\
    1, & \text{otherwise}
    \end{cases}
\end{equation}
Specifically, any step that fails syntax checks or duplicates history is deemed invalid and strictly penalized with a zero score.


\paragraph{Recall-based Soft Reward}
To quantify the information gain of intermediate steps, we introduce an execution-based soft reward $r_{\text{soft}}$.
We posit that the multi-step reasoning process parallels SQL query execution, which proceeds from broad filtering to specific refinement.
Consequently, intermediate execution results $E(a_t)$ naturally serve as {supersets} of the ground truth $E(y)$.
Therefore, we employ {cell-level recall} to define $r_{\text{soft}}$:
\begin{equation}
    r_{\text{soft}}(a_t) = \frac{| E(a_t) \cap E(y) |}{| E(y) | + \epsilon}
\end{equation}
where $E(\cdot)$ denotes the flattened set of cell values.


\paragraph{Comprehensive Process Reward}
The final process reward $r_t^{\text{proc}}$ combines the hard constraint and soft reward.
Crucially, the soft reward is activated only if the hard constraint is met:
\begin{equation}
    r_t^{\text{proc}} = I_{\text{hard}}(a_t) \cdot \left( 1 + r_{\text{soft}}(a_t) \right)
\end{equation}



\subsection{Step-level Credit Assignment}
Despite the rich signals from multi-granular rewards, direct optimization still faces challenges: 
(1) How to attribute credit to individual steps for the outcome reward?
(2) How to incentivize indispensable "exploration actions" that yield low immediate returns?
(3) How to distinguish high-quality steps from redundant operations, overcoming the limitations of trajectory-level normalization?
To address these, we propose a framework comprising Outcome Reward Discounting, Process Reward Smoothing, and Step-level Advantage Estimation.


\subsubsection{Outcome Reward Discounting}
In multi-step reasoning, actions exhibit temporal heterogeneity: early steps lay the foundation, whereas late steps directly determine the outcome. Consequently, uniformly distributing the final reward $R^{\text{out}}$ results in ambiguous credit assignment.
To differentiate these contributions, we introduce a discount factor $\gamma$ to propagate the reward backward, yielding a time-decayed value estimate $V^{\text{out}}_t$:
\begin{equation}
    V^{\text{out}}_t = \gamma^{T-t} \cdot R^{\text{out}}, t \in [0, T],\gamma \in (0, 1)
\end{equation}
This decay enforces temporal proximity, prioritizing recent causal actions while sustaining incentives for early decisions, to ensure more discriminative credit assignment.


\subsubsection{Process Reward Smoothing}
Rule-based process rewards may suffer from low recall, causing critical exploration actions to be overlooked. This can mislead the model into evading these essential steps and falling into local optima. 
To capture the underlying logical dependencies, we propose an {reverse smoothing mechanism} that backtracks future potential to the current step. 
The smoothed process value $V^{\text{proc}}_t$ is defined as:

{
\small
\setlength{\abovedisplayskip}{0pt} 
\setlength{\belowdisplayskip}{5pt} 
\begin{equation}
    V^{\text{proc}}_t = 
    \begin{cases} 
    r^{\text{proc}}_t, & \text{if } t = T -1 \\
    (1-\beta) \cdot r^{\text{proc}}_t + \beta \cdot V^{\text{proc}}_{t+1}, & \text{if } t \in [0, T-1)
    \end{cases}
\end{equation}
}

where $\beta \in [0, 1]$ is the smoothing coefficient. By integrating subsequent value, this mechanism affirms the instrumental value of exploration, encouraging the model to prioritize global planning over myopic optimization.

\subsubsection{Step-level Advantage Estimation}


Given the global signal $V^{\text{out}}$ and the local signal $V^{\text{proc}}$, the key challenge is to combine them into a unified advantage metric with discriminative step-level credit assignment.
To address this, we perform group-level normalization over all valid steps across the sampled trajectories.
Specifically, for a group of $G$ sampled trajectories where trajectory $i$ has $T_i$ valid steps, we construct value sets $\mathcal{V}^{\text{out}} = \{V^{\text{out}}_{i,t} \mid i=1,\ldots,G,\; t=0,\ldots,T_i\}$ and $\mathcal{V}^{\text{proc}} = \{V^{\text{proc}}_{i,t} \mid i=1,\ldots,G,\; t=0,\ldots,T_i\}$ collecting values from all valid steps regardless of their position.
The standardized advantages are then computed using the mean and standard deviation over these sets:

{\small
\setlength{\abovedisplayskip}{0pt} 
\setlength{\belowdisplayskip}{5pt} 
\begin{equation}
\hat{A}^{\text{out}}_{i,t} =
\frac{V^{\text{out}}_{i,t} - \text{Mean}(\mathcal{V}^{\text{out}})}
{\text{Std}(\mathcal{V}^{\text{out}}) + \epsilon}, 
\quad
\hat{A}^{\text{proc}}_{i,t} =
\frac{V^{\text{proc}}_{i,t} - \text{Mean}(\mathcal{V}^{\text{proc}})}
{\text{Std}(\mathcal{V}^{\text{proc}}) + \epsilon}.
\end{equation}
}
The final step-level advantage is computed as

{\small
\setlength{\abovedisplayskip}{0pt} 
\setlength{\belowdisplayskip}{5pt} 
\begin{equation}
A_{i,t} =
\begin{cases}
\hat{A}^{\text{out}}_{i,t}, & t = T \\
\lambda \cdot \hat{A}^{\text{out}}_{i,t} + (1-\lambda) \cdot \hat{A}^{\text{proc}}_{i,t}, & \text{otherwise}.
\end{cases}
\end{equation}
}

The advantage $A_{i,t}$ is broadcast to all tokens ($r_t$, $a_t$, and $e_t$) of step $t$ in trajectory $i$. 
Since the final step $t=T$ generates the answer, only the outcome signal is used.
This design removes baseline discrepancies and enables precise reinforcement of correct intermediate reasoning steps, even in trajectories that ultimately fail.

\subsection{Policy Optimization}

Following the GRPO~\citep{GRPO} paradigm, we optimize the policy $\pi_\theta$ by maximizing a surrogate objective over the sampled group. 
Unlike trajectory-level GRPO, we use the step-level advantage $A_{i,t}$ for finer credit assignment.
The objective is defined as:

{\small 
\setlength{\abovedisplayskip}{0pt} 
\setlength{\belowdisplayskip}{5pt} 
\begin{multline} 
\mathcal{J}(\theta) = \mathbb{E}_{x \sim \mathcal{D}, \{y_i\}_{i=1}^G \sim \pi_{\theta_{\text{old}}}(\cdot|x;\mathcal{E})} \\
\left[
\frac{1}{G} \sum_{i=1}^G \frac{1}{T_i} \sum_{t=1}^{T_i} 
\left(
\min\left( 
r_{i,t}(\theta) \cdot A_{i,t}, 
{clip}\right)
- \beta \, \mathbb{D}_{\text{KL}}
\right)
\right], \\
{clip} = \text{clip}\left( {r_{i,t}(\theta)},
1-\epsilon, 
1+\epsilon \right) A_{i,t}
\end{multline} }

where $r_{i,t}(\theta) = \frac{\pi_\theta(a_{i,t}|s_{i,t})}{\pi_{\theta_{\text{old}}}(a_{i,t}|s_{i,t})}$ is the probability ratio, and $\beta$ is the KL penalty coefficient. 
Crucially, we mask environment feedback tokens during backpropagation, ensuring the optimization targets only the model's generative logic (thoughts and actions) rather than tool outputs.

\section{Experiments}


\begin{table*}[t]
\centering
\footnotesize 
\setlength{\tabcolsep}{2pt} 
\begin{tabular}{lcccccccccccc}
\toprule
\multirow{2}{*}{\textbf{Method}} & \multicolumn{2}{c}{BIRD Dev} & \multicolumn{2}{c}{Spider Dev} & \multicolumn{1}{c}{Spider Test} & \multicolumn{2}{c}{Spider-Realistic} & \multicolumn{2}{c}{Spider-Syn} & \multicolumn{1}{c}{Spider-DK} & \multirow{2}{*}{\textbf{Avg.}} \\
\cmidrule(r){2-3} \cmidrule(lr){4-5} \cmidrule(lr){6-6} \cmidrule(lr){7-8} \cmidrule(lr){9-10} \cmidrule(l){11-11} 
& EX & VES & EX & TS & EX  & EX  & TS & EX  & TS  & EX & \\
\midrule
\multicolumn{12}{l}{\textit{Model Size: 4B}} \\
No-Train & 65.58 & 67.97 & 85.6 & 75.4 & 85.2 & 82.5 & 66.3 & 76.1 & 65.0 & 73.6 & 74.34 \\
GRPO     & 66.23 & \color{gray}\textbf{70.65} & 88.5 & 81.4 & 85.0 & 83.7 & 75.2 & 78.9 & 71.2 & \color{gray}\textbf{80.7} & 78.15 \\
DAPO     & 67.67 & 69.93 & \color{gray}\textbf{89.7} & \color{gray}\textbf{84.1} & \color{gray}\textbf{89.3} & 87.2 & 78.5 & 80.0 & \color{gray}\textbf{72.5} & 79.6 & 79.85 \\
GSPO     & \color{gray}\textbf{68.06} & 69.99 & 89.5 & 83.9 & 89.2 & 86.0 & 78.1 & 80.1 & 72.1 & 78.9 & 79.59 \\
GIGPO    & 67.99 & 70.20 & 89.6 & 83.4 & 89.1 & \color{gray}\textbf{87.6} & \color{gray}\textbf{80.5} & \color{gray}\textbf{80.2} & 71.7 & 80.4 & \color{gray}\textbf{80.07} \\
\textbf{FineStep (Our)} & \textbf{68.51}\textsubscript{$\uparrow$2.28} & \textbf{71.65}\textsubscript{$\uparrow$1.00} & \textbf{90.3}\textsubscript{$\uparrow$1.8} & \textbf{84.3}\textsubscript{$\uparrow$2.9} & \textbf{89.9}\textsubscript{$\uparrow$4.9} & \textbf{88.6}\textsubscript{$\uparrow$4.9} & \textbf{80.9}\textsubscript{$\uparrow$5.7} & \textbf{83.1}\textsubscript{$\uparrow$4.2} & \textbf{75.2}\textsubscript{$\uparrow$4.0} & \textbf{81.5}\textsubscript{$\uparrow$0.8} & \textbf{81.40}\textsubscript{$\uparrow$3.25} \\
\midrule
\multicolumn{12}{l}{\textit{Model Size: 8B}} \\
No-Train & 62.71 & 66.76 & 84.4 & 72.1 & 85.0 & 80.9 & 65.4 & 75.4 & 62.0 & 75.0 & 72.97 \\
GRPO     & 66.56 & 68.13 & 87.2 & 79.1 & 88.7 & 83.7 & 75.8 & 79.2 & 70.0 & 78.7 & 77.71 \\
DAPO     & 65.97 & 68.36 & \color{gray}\textbf{88.6} & 81.2 & 88.5 & 85.4 & 76.0 & 79.5 & 71.1 & 79.4 & 78.40 \\
GSPO     & 66.62 & \color{gray}\textbf{69.40} & 88.4 & \color{gray}\textbf{81.5} & \color{gray}\textbf{89.5} & 84.8 & 77.0 & 79.6 & 70.3 & \textbf{80.2} & 78.73 \\
GIGPO    & \color{gray}\textbf{66.88} & 68.27 & \color{gray}\textbf{88.6} & 81.2 & 89.4 & \color{gray}\textbf{85.6} & \color{gray}\textbf{77.0} & \color{gray}\textbf{80.4} & \textbf{71.7} & 79.1 & \color{gray}\textbf{78.82} \\
\textbf{FineStep (Our)} & \textbf{68.06}\textsubscript{$\uparrow$1.50} & \textbf{70.90}\textsubscript{$\uparrow$2.77} & \textbf{88.9}\textsubscript{$\uparrow$1.7} & \textbf{82.4}\textsubscript{$\uparrow$3.3} & \textbf{89.9}\textsubscript{$\uparrow$1.2} & \textbf{86.0}\textsubscript{$\uparrow$2.3} & \textbf{77.9}\textsubscript{$\uparrow$2.1} & \textbf{81.4}\textsubscript{$\uparrow$2.2} & \color{gray}\textbf{71.6}\textsubscript{$\uparrow$1.6} & \color{gray}\textbf{80.0}\textsubscript{$\uparrow$1.3} & \textbf{79.71}\textsubscript{$\uparrow$2.00} \\
\midrule
\multicolumn{12}{l}{\textit{Model Size: 30B-A3B}} \\
No-Train & 67.08 & 69.63 & 82.4 & 69.2 & 82.1 & 82.5 & 64.6 & 75.4 & 61.2 & 75.3 & 72.94 \\
GRPO     & \color{gray}\textbf{68.32} & \color{gray}\textbf{70.77} & \textbf{88.0} & \color{gray}\textbf{78.6} & \color{gray}\textbf{88.9} & \color{gray}\textbf{83.3} & \color{gray}\textbf{73.6} & \color{gray}\textbf{79.6} & \color{gray}\textbf{70.4} & \color{gray}\textbf{78.7} & \color{gray}\textbf{78.02} \\
\textbf{FineStep (Our)} & \textbf{69.23}\textsubscript{$\uparrow$0.91} & \textbf{73.30}\textsubscript{$\uparrow$2.53} & \color{gray}\textbf{87.6}\textsubscript{$\downarrow$0.4} & \textbf{81.2}\textsubscript{$\uparrow$2.6} & \textbf{89.5}\textsubscript{$\uparrow$0.6} & \textbf{84.1}\textsubscript{$\uparrow$0.8} & \textbf{76.0}\textsubscript{$\uparrow$2.4} & \textbf{80.8}\textsubscript{$\uparrow$1.2} & \textbf{72.6}\textsubscript{$\uparrow$2.2} & \textbf{81.9}\textsubscript{$\uparrow$3.2} & \textbf{79.62}\textsubscript{$\uparrow$1.60} \\
\bottomrule
\end{tabular}
\caption{Overall performance of different algorithms. 
Subscripts indicate improvements over GRPO. 
The best results are highlighted in \textbf{bold}, and the second-best in {\color{gray}\textbf{gray bold}}. 
The same notation is used in Tables~\ref{tab:exp24_bird_decoding_comparison} and~\ref{tab:exp25difficulty_comparison}.}
\label{tab:exp1_main_results_vs_grpo}
\end{table*}

\paragraph{Datasets and Metrics}
We evaluate our method on two widely used text-to-SQL benchmarks: BIRD~\citep{DBLP:conf/nips/bird} and Spider~\citep{DBLP:conf/emnlp/spider}. 
For BIRD, we report execution accuracy (EX) and valid efficiency score (VES), where VES measures the execution efficiency of predicted SQL queries relative to the ground truth. 
For Spider, we use execution accuracy (EX) and test-suite accuracy (TS)~\cite{Test-Suites}, which verifies whether predicted SQL queries produce correct results across multiple database test cases. 
We also evaluate three Spider robustness variants: Spider-Syn~\citep{DBLP:conf/acl/Spider-Syn}, Spider-Realistic~\citep{DBLP:conf/naacl/Spider-Realistic}, and Spider-DK~\citep{DBLP:conf/emnlp/Spider-DK}. 
More details are provided in Appendix~\ref{app:dataset}.

\paragraph{Baselines}
We compare our method with four RL algorithms: GRPO~\citep{GRPO}, DAPO~\citep{dapo}, GSPO~\citep{gspo}, and GIGPO~\citep{gigpo}. 
Among them, GIGPO clusters intermediate reasoning trajectories to normalize process-level advantages, while the others compute advantages solely from outcome rewards.

\paragraph{Implementation Details}
We evaluate three models: Qwen3-4B-Instruction, Qwen3-8B, and Qwen3-30B-A3B, representing an instruction model, a base model, and an MoE model, respectively. 
All RL algorithms are implemented in the Verl~\citep{verl} framework with a batch size of 128, 5 epochs, and a group size of 8. 
Database execution tools are implemented using SQLite~\citep{sqlite} and FastAPI~\citep{fastapi}. 
Unless otherwise specified, all results are obtained by sampling 8 responses with temperature $T=1$ and selecting the final prediction via self-consistency. 
For FineStep, the smoothing coefficient $\beta$ and advantage weight $\lambda$ are both set to 0.5, discount factor $\gamma$ is set to 0.98.
More implementation details are provided in Appendix~\ref{app:exp_set}.

\subsection{Overall Performance} \label{exp:Overall}

\paragraph{Cross-Model Adaptability}
Table~\ref{tab:exp1_main_results_vs_grpo} reports the performance of different RL algorithms on the BIRD and Spider benchmarks across three model configurations (4B, 8B, and 30B-A3B). 
FineStep consistently achieves the best average performance across all model architectures. 
Compared with GRPO, FineStep improves the average score by +3.25\%, +2.00\%, and +1.60\% on the 4B, 8B, and 30B-A3B models, respectively. 
These results demonstrate that FineStep adapts well to different model scales and architectures, including instruction-tuned, base, and MoE models.

\paragraph{Comparison with Baselines}
FineStep consistently outperforms existing RL methods (GRPO, DAPO, GSPO, and GIGPO) on most evaluation metrics. 
By introducing step-level credit assignment, FineStep provides denser supervision for long-chain SQL reasoning. This design mitigates cases where correct final execution masks flawed intermediate reasoning, leading to reliable result.

\begin{figure*}[t]
    \includegraphics[width=1\textwidth]{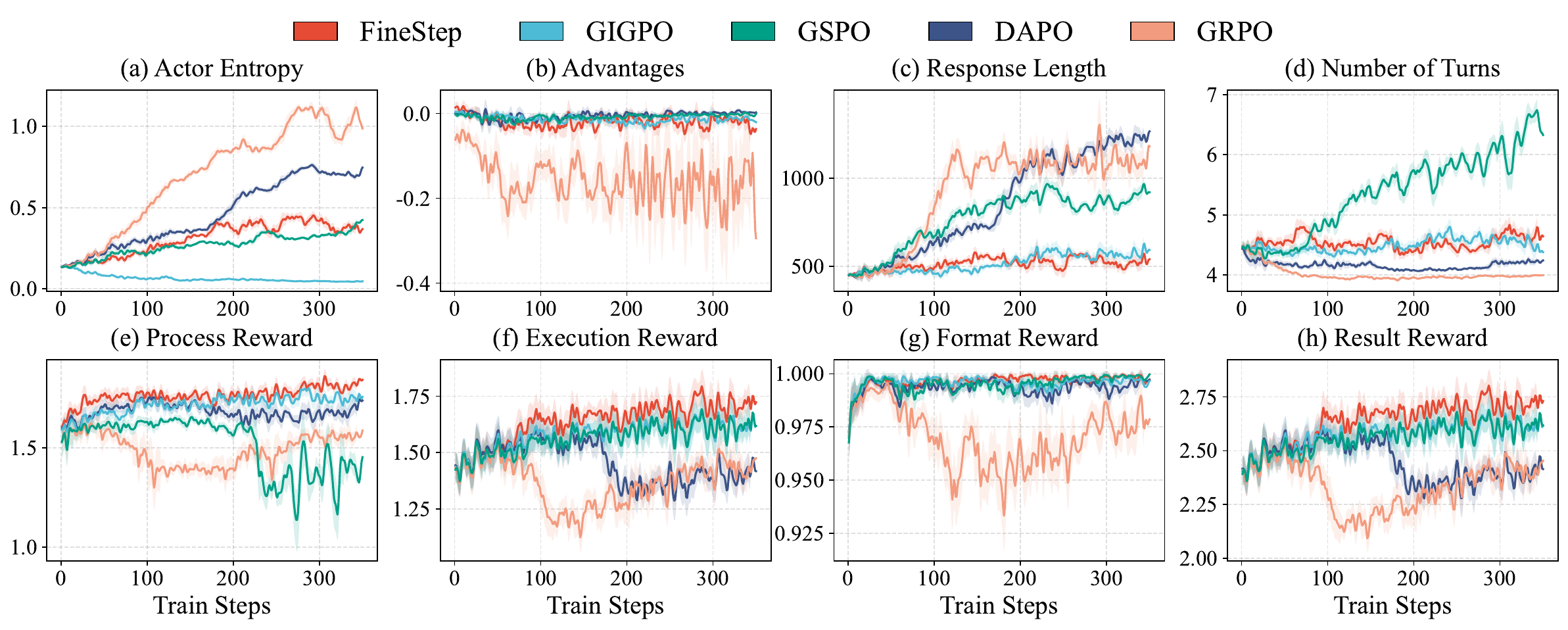}
    \caption{Comparison of indicators of different algorithms during RL training.}
    \label{fig:exp22_ablation_of_components}
\end{figure*}

\begin{figure}[t]
    \includegraphics[width=1\columnwidth]{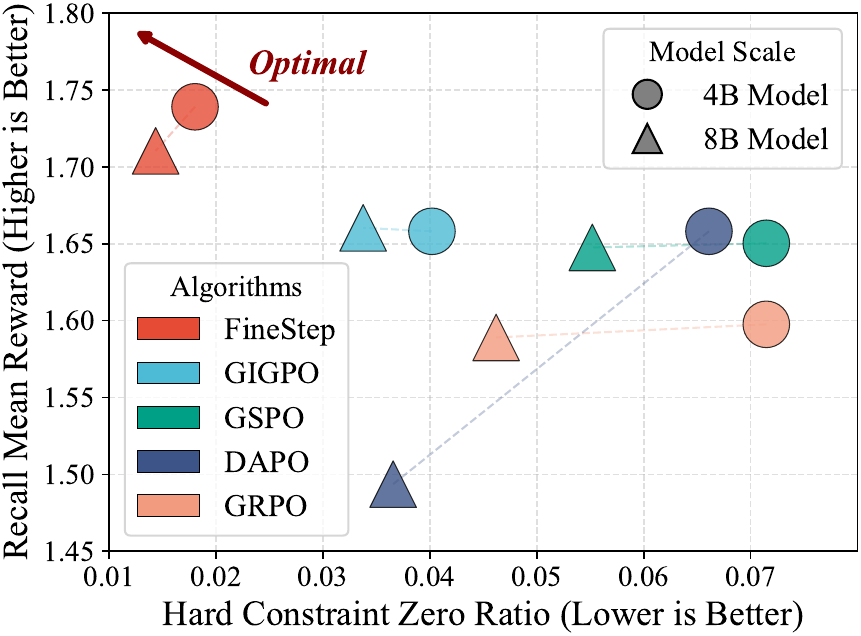}
    \caption{Comparison of process rewards on Bird Dev.}
    \label{fig:exp21_reward_compare}
\end{figure}

\subsection{Process Reward Behavior}

Figure~\ref{fig:exp21_reward_compare} shows the relationship between the hard constraint zero ratio and the recall-based process reward ($r_{\text{soft}}$) across different RL algorithms. 
FineStep consistently occupies the upper-left region of the plot across both model scales (4B and 8B), achieving the lowest violation rates (0.018 and 0.014) and the highest recall-based rewards (1.739 and 1.710), indicating more reliable tool usage and more informative intermediate execution. 
In contrast, other RL methods mostly fall into the lower-left region with higher violation rates and lower process rewards, with GRPO performing the worst.
GIGPO performs better than other baselines due to step-level advantage normalization, but still trails FineStep with its explicit step-level reward and advantage design.
These results show that outcome-only optimization cannot ensure valid and informative reasoning trajectories, while FineStep shifts the trajectory distribution toward more reliable paths.

\subsection{Training Dynamics Analysis}
Figure~\ref{fig:exp22_ablation_of_components} illustrates the training dynamics of different RL algorithms. 
FineStep exhibits more stable optimization and consistently higher reward signals. 
As shown in Fig.~\ref{fig:exp22_ablation_of_components}(a), FineStep maintains moderate actor entropy, avoiding the entropy collapse in GSPO and the entropy explosion in GRPO. 
In terms of reasoning efficiency, Fig.~\ref{fig:exp22_ablation_of_components}(c,d) shows that FineStep produces more concise trajectories with fewer interaction turns, whereas GSPO tends to generate longer responses with redundant tool calls. 
FineStep also achieves higher process and execution rewards with near-perfect formatting (Fig.~\ref{fig:exp22_ablation_of_components}(e,f,g,h)), while baselines such as DAPO and GRPO suffer reward collapse in later training stages. 
Overall, these results indicate that FineStep shifts the reasoning trajectory distribution toward more reliable and informative paths.

\begin{table}[t]
\centering
\small
\setlength{\tabcolsep}{8pt} 
\begin{tabular}{lcc}
\toprule
{Configuration on Bird Dev} & \textbf{EX (\%)} & \textbf{VES (\%)} \\
\midrule
GRPO (Baseline) & 66.56 & 68.13 \\
\hdashline[1pt/2pt] \noalign{\smallskip}
FineStep & & \\
\quad w/ \textit{Precision-based Reward} & 66.95 & 69.24 \\
\quad w/ \textit{F1-based Reward}        & 66.82 & 69.28 \\
\quad w/ \textit{Recall-based Reward}    & \textbf{68.06} & \textbf{70.90} \\
\bottomrule
\end{tabular}
\caption{Comparison of soft reward on FineStep-8B.}
\label{tab:exp23soft_reward_comparison}
\end{table}

\subsection{Effect of Soft Reward Design}
Table~\ref{tab:exp23soft_reward_comparison} shows that the recall-based soft reward achieves the best performance among all variants.
This advantage stems from the nature of multi-step Text-to-SQL reasoning, where intermediate execution results often form supersets of the final answer. 
In such cases, recall better captures whether useful information is preserved during intermediate steps. 
By contrast, precision-based and F1-based rewards tend to penalize exploratory yet informative intermediate results.

\begin{table}[t]
\centering
\small
\begin{tabular}{lccc}
\toprule
\multirow{2}{*}{\textbf{Method}} & \multicolumn{3}{c}{\textbf{BIRD Dev (\%)}} \\
\cmidrule(lr){2-4}
& \textbf{Greedy} & \textbf{Voting@8} & \textbf{Pass@8} \\
\midrule
\multicolumn{4}{l}{\textit{Model Size: 4B}} \\
Training-free & 63.95 & 65.58 & 73.53 \\
GRPO          & 65.78 & 66.23 & 71.77 \\
DAPO          & 66.75 & 67.67 & 72.29 \\
GSPO          & \textbf{67.01} & \color{gray}\textbf{68.06} & \color{gray}\textbf{74.05} \\
GIGPO         & \color{gray}\textbf{66.82} & 67.99 & 73.08 \\
\textbf{FineStep (Our)} & \textbf{67.01}\textsubscript{$\uparrow$1.23} & \textbf{68.51}\textsubscript{$\uparrow$2.28} & \textbf{74.58}\textsubscript{$\uparrow$2.81} \\
\midrule
\multicolumn{4}{l}{\textit{Model Size: 8B}} \\
Training-free & 56.13 & 62.71 & 69.43 \\
GRPO          & 64.60 & 66.56 & 73.73 \\
DAPO          & 63.10 & 65.97 & 73.21 \\
GSPO          & 64.86 & 66.62 & \color{gray}\textbf{73.79} \\
GIGPO         & \color{gray}\textbf{65.19} & \color{gray}\textbf{66.88} & 73.53 \\
\textbf{FineStep (Our)} & \textbf{65.78}\textsubscript{$\uparrow$1.18} & \textbf{68.06}\textsubscript{$\uparrow$1.50} & \textbf{74.64}\textsubscript{$\uparrow$0.91} \\
\bottomrule
\end{tabular}
\caption{Comparison of decoding strategies.}
\label{tab:exp24_bird_decoding_comparison}
\end{table}

\subsection{Decoding Strategy Comparison}
Table~\ref{tab:exp24_bird_decoding_comparison} compares decoding strategies on the BIRD Dev.
FineStep achieves the best performance across all settings and model scales.
Under greedy decoding, it matches or surpasses the strongest baselines, indicating improved policy quality.
With multiple samples, FineStep further improves Voting@8 and Pass@8 (e.g., +2.28\% and +2.81\% on 4B over GRPO).
This suggests that FineStep produces a higher-quality candidate distribution, allowing self-consistency to more reliably select correct SQL queries.

\begin{table}[t]
\small
\setlength{\tabcolsep}{2pt} 
\centering
\begin{tabular}{lcccc}
\toprule
\textbf{Method} & \textbf{Easy} & \textbf{Mid} & \textbf{Hard} & \textbf{Total} \\
\midrule
\multicolumn{5}{l}{\textit{Model Size: 4B}} \\
No-Train      & 71.24 & 59.48 & 48.97 & 65.58 \\
GRPO          & 71.68 & 59.48 & 53.10 & 66.23 \\
DAPO          & 73.73 & 59.05 & 56.55 & 67.67 \\
GSPO          & 73.19 & \textbf{60.99} & 57.93 & \color{gray}\textbf{68.06} \\
GIGPO         & \color{gray}\textbf{73.95} & 59.48 & \color{gray}\textbf{57.24} & 67.99 \\
\textbf{FineStep (Our)} & \textbf{74.16}\textsubscript{$\uparrow$2.48} & \color{gray}\textbf{60.13}\textsubscript{$\uparrow$0.65} & \textbf{59.31}\textsubscript{$\uparrow$6.21} & \textbf{68.51}\textsubscript{$\uparrow$2.28} \\
\midrule
\multicolumn{5}{l}{\textit{Model Size: 8B}} \\
No-Train      & 70.27 & 54.09 & 42.07 & 62.71 \\
GRPO          & \color{gray}\textbf{73.41} & 58.19 & 49.66 & 66.56 \\
DAPO          & 72.86 & 56.90 & 51.03 & 65.97 \\
GSPO          & 73.19 & 57.76 & \color{gray}\textbf{53.10} & 66.62 \\
GIGPO         & \color{gray}\textbf{73.41} & \color{gray}\textbf{58.84} & 51.03 & \color{gray}\textbf{66.88} \\
\textbf{FineStep (Our)} & \textbf{74.05}\textsubscript{$\uparrow$0.64} & \textbf{58.84}\textsubscript{$\uparrow$0.65} & \textbf{59.31}\textsubscript{$\uparrow$9.65} & \textbf{68.06}\textsubscript{$\uparrow$1.50} \\
\bottomrule
\end{tabular}
\caption{Difficulty analysis on Bird Dev.}
\label{tab:exp25difficulty_comparison}
\end{table}

\subsection{Performance Across Difficulty Levels}
Table~\ref{tab:exp25difficulty_comparison} reports EX score across different query difficulty levels on the BIRD Dev.
FineStep consistently achieves the best overall performance across all model sizes. The improvements are most pronounced on hard queries, with gains of +6.21\% and +9.65\% on the 4B and 8B models over GRPO.
This suggests that FineStep benefits complex reasoning scenarios, where multi-step SQL generation requires reliable intermediate decisions.

\begin{table}[t]
\small
\centering
\setlength{\tabcolsep}{2pt} 
\begin{tabular}{lccc}
\toprule
\textbf{Method / Ablation} & \textbf{Greedy} & \textbf{Voting@8} & \textbf{Pass@8} \\
\midrule
FineStep                       & \textbf{67.01} & \textbf{68.51} & \textbf{74.58} \\
\hdashline[1pt/2pt] \noalign{\smallskip}
\hspace{0.5em} w/o Hard Reward $I_{\text{hard}}$   & 66.95\textsubscript{$\downarrow$0.06} & 67.41\textsubscript{$\downarrow$1.10} & 73.86\textsubscript{$\downarrow$0.72} \\
\hspace{0.5em} w/o Soft Reward $r_{\text{soft}}$  & 66.17\textsubscript{$\downarrow$0.84} & 67.21\textsubscript{$\downarrow$1.30} & 72.69\textsubscript{$\downarrow$1.89} \\
\hdashline[1pt/2pt] \noalign{\smallskip}
\hspace{0.5em} No discounting ($\gamma=1$) & 62.67\textsubscript{$\downarrow$4.34} & 66.62\textsubscript{$\downarrow$1.89} & 73.01\textsubscript{$\downarrow$1.57} \\
\hspace{0.5em} No smoothing ($\beta=0$)  & 64.93\textsubscript{$\downarrow$2.08} & 65.84\textsubscript{$\downarrow$2.67} & 72.75\textsubscript{$\downarrow$1.83} \\
\hdashline[1pt/2pt] \noalign{\smallskip}
\hspace{0.5em} Only $\hat{A}^{\text{out}}_{i,t}$ ($\lambda=1$)     & 65.19\textsubscript{$\downarrow$1.82} & 65.97\textsubscript{$\downarrow$2.54} & 72.75\textsubscript{$\downarrow$1.83} \\
\hspace{0.5em} Only $\hat{A}^{\text{proc}}_{i,t}$ ($\lambda=0$)        & 66.43\textsubscript{$\downarrow$0.58} & 67.41\textsubscript{$\downarrow$1.10} & 73.21\textsubscript{$\downarrow$1.37} \\
\bottomrule
\end{tabular}
\caption{Ablation of FineStep-4B on BIRD Dev.}
\label{tab:exp31_ablation_of_components}
\end{table}





\subsection{Ablation Study}
Table~\ref{tab:exp31_ablation_of_components} presents an ablation study of FineStep, examining three components: reward design, reward propagation, and advantage composition.

\paragraph{Effect of Reward Design.}
Removing either the hard constraint reward $I_{\text{hard}}$ or the soft reward $r_{\text{soft}}$ degrades performance. 
The larger drop without $r_{\text{soft}}$ suggests that recall-based process reward provides the primary learning signal for intermediate reasoning, while $I_{\text{hard}}$ mainly stabilizes tool usage by preventing invalid or repetitive actions.

\paragraph{Effect of Reward Propagation.}
Removing outcome reward discounting ($\gamma=1$) or process reward smoothing ($\beta=0$) reduces performance, indicating that both mechanisms help stabilize credit assignment across long reasoning trajectories.

\paragraph{Effect of Advantage Composition.}
Using only outcome advantage ($\lambda=1$) or only process advantage ($\lambda=0$) both harms performance, showing that outcome supervision and process supervision are complementary: the former provides global task-level guidance, while the latter supplies fine-grained signals for intermediate reasoning steps.

\begin{figure}[t]
    \includegraphics[width=1\columnwidth]{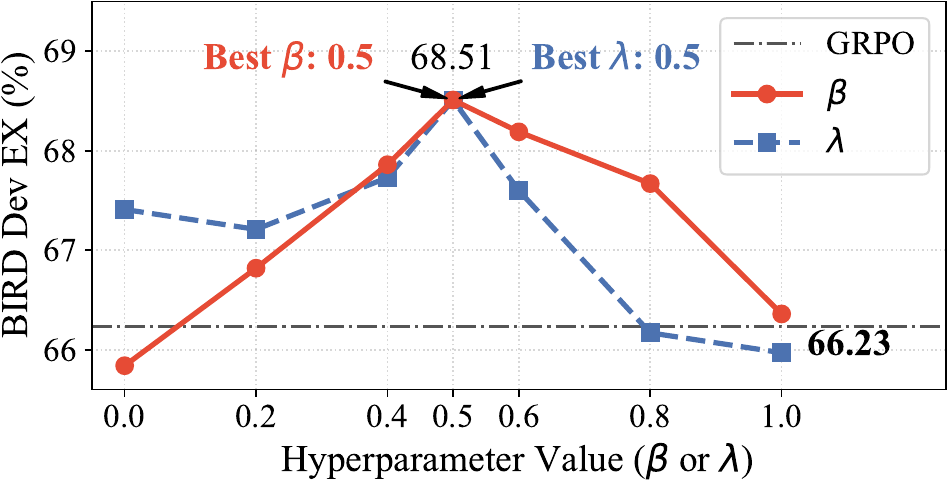}
    \caption{Sensitivity analysis of FineStep-4B.}
    \label{fig:exp32_ablation_of_parameter}
\end{figure}

\subsection{Hyperparameter Sensitivity}
Figure~\ref{fig:exp32_ablation_of_parameter} analyzes the effects of the smoothing coefficient $\beta$ and the advantage mixing weight $\lambda$.
Performance peaks at $\beta=0.5$ and $\lambda=0.5$, indicating that both moderate reward smoothing and balanced process–outcome advantages are important, while extreme values lead to noticeable performance degradation.
Across a wide range of hyperparameter values, FineStep consistently outperforms GRPO, demonstrating that the improvements mainly stem from the joint design of process and outcome supervision. More supplementary information is in Appendix~\ref{app:hyperparam}.

\begin{figure}[t]
    \includegraphics[width=1\columnwidth]{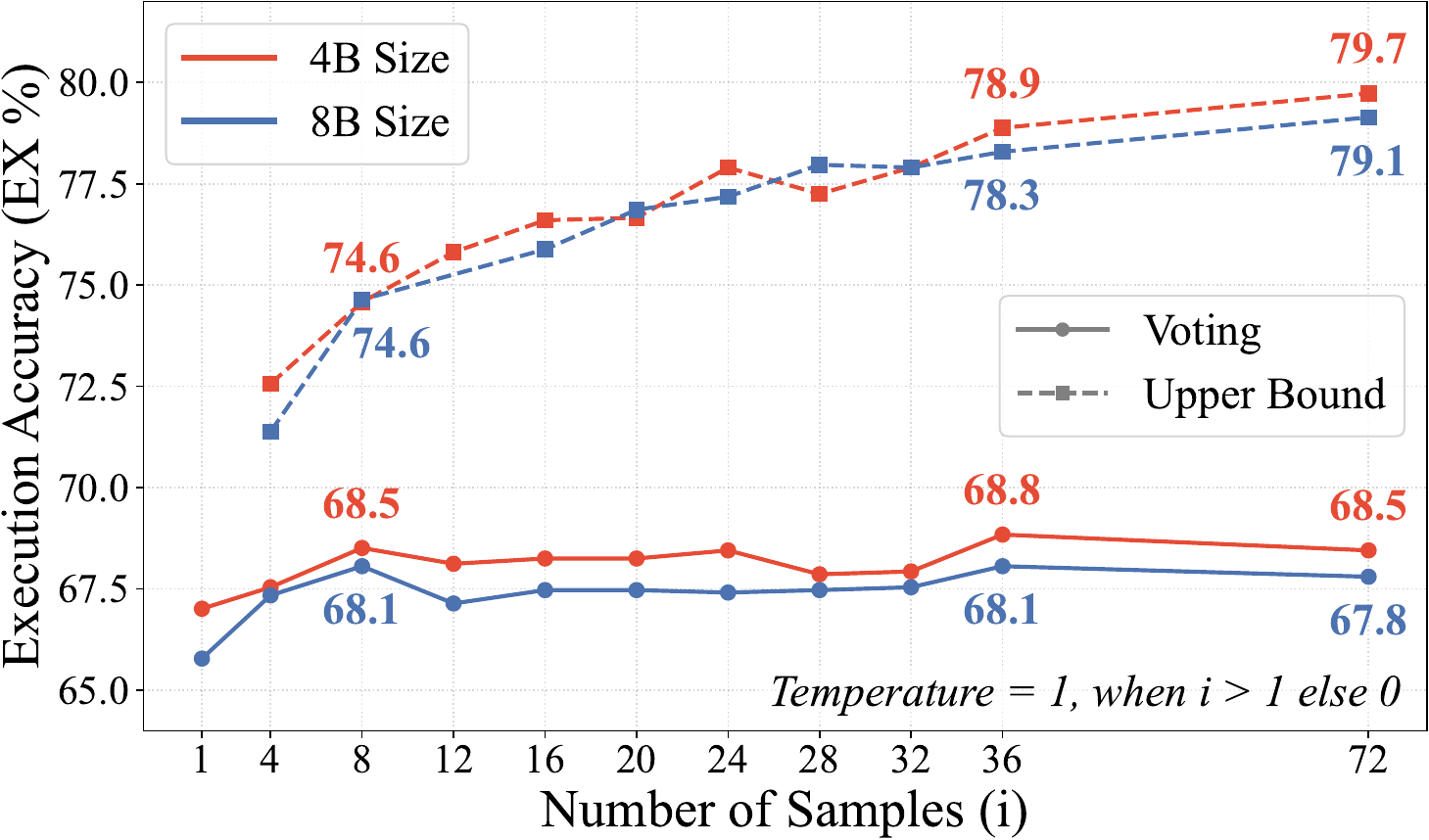}
    \caption{The impact of sample numbers on Bird Dev.}
    \label{fig:exp35_sample_num}
\end{figure}

\subsection{Effect of Sampling Size}
Figure~\ref{fig:exp35_sample_num} analyzes the effect of inference sample size on execution accuracy.
As the number of samples increases, the upper bound (Pass@k) steadily improves for both 4B and 8B models, indicating higher chances of generating correct SQL queries. 
However, self-consistency voting quickly saturates.
For example, increasing the sample size from 8 to 72 raises the upper bound from 74.6\% to 79.7\% on the 4B model, while voting accuracy only increases from 68.5\% to 68.8\%.

\section{Conclusion} \label{conclusion}

In this work, we address the credit assignment challenge in tool-augmented Text-to-SQL reasoning, where existing RL methods rely on outcome-level rewards and fail to distinguish effective intermediate steps from redundant or erroneous actions. 
To address this limitation, we propose \textbf{FineStep}, a fine-grained credit assignment framework that augments outcome supervision with independent process rewards and introduces a multi-dimensional step-level advantage estimation strategy.
Extensive experiments on benchmarks (BIRD and Spider) show that FineStep consistently improves execution accuracy across different model architectures and decoding settings. 
Further analyses demonstrate that FineStep stabilizes RL training, reduces invalid tool interactions, and produces more concise and reliable reasoning trajectories.
Overall, our results highlight the importance of explicitly modeling intermediate reasoning steps for improving long-horizon agentic reasoning, and suggest that fine-grained credit assignment is key to building reliable tool-augmented reasoning systems.

\section*{Limitation} \label{limitation}

While FineStep demonstrates strong improvements for Text-to-SQL through process-based reinforcement learning, several limitations remain. First, the current framework is tailored to the structured characteristics of SQL, where step-level decomposition and reward signals can be explicitly defined. Extending FineStep to broader reasoning domains—such as mathematical problem solving or multi-hop code generation—remains an important direction for future work. Second, the current reward design mainly relies on rule-based signals derived from predefined heuristics and execution results. Although this ensures high precision and interpretability, it may lack flexibility when handling ambiguous or unconventional reasoning paths. In future work, we plan to incorporate a learned Process-supervised Reward Model (PRM), trained with human feedback or high-quality model distillations, to provide more nuanced step-level evaluations and improve robustness in complex reasoning scenarios.

\bibliography{custom}

\newpage

\appendix

\section{Appendices}

\subsection{Benchmarks}\label{app:dataset}

In this study, we evaluate our method on several representative Text-to-SQL benchmarks.

\textbf{BIRD}~\citep{DBLP:conf/nips/bird} is a large-scale cross-domain Text-to-SQL benchmark designed for realistic applications. 
It contains 12,751 natural language questions paired with SQL queries across 95 databases from 37 domains, with a total data scale of 33.4 GB. 
Compared with earlier datasets such as Spider~\citep{DBLP:conf/emnlp/spider} and WikiSQL~\citep{DBLP:WikiSQL}, BIRD places greater emphasis on database contents (e.g., field values and real data distributions), making it closer to real-world query scenarios.

\textbf{Spider}~\citep{DBLP:conf/emnlp/spider} is a widely used Text-to-SQL benchmark consisting of 8,659 training examples, 1,034 development examples, and 2,147 test examples across 200 databases from 138 domains. 
It adopts a cross-database evaluation setting, where models must generalize to unseen database schemas.

\textbf{Spider-DK}~\citep{DBLP:conf/emnlp/Spider-DK}, \textbf{Spider-Syn}~\citep{DBLP:conf/acl/Spider-Syn}, and \textbf{Spider-Realistic}~\citep{DBLP:conf/naacl/Spider-Realistic} are three variants derived from Spider to better simulate real-world queries. 
Spider-DK introduces domain knowledge and semantic distractions to increase reasoning difficulty, Spider-Syn evaluates robustness to paraphrasing and syntactic variations, and Spider-Realistic reformulates queries into more conversational and incomplete forms.

Table~\ref{table:dataset} summarizes the statistics of the Spider and BIRD datasets. 
In general, SQL queries in BIRD are more complex than those in Spider, reflecting its closer alignment with real-world database applications.

\subsection{Reinforcement Learning Baselines} \label{app:baseline}

This appendix briefly introduces the reinforcement learning (RL) baseline algorithms used in this study. 
All selected methods follow a critic-free or group-relative optimization paradigm, which removes the need for an auxiliary value network and improves training efficiency for large reasoning models.

\subsubsection{Group Relative Policy Optimization}

Group Relative Policy Optimization (GRPO)~\citep{GRPO}, introduced in the DeepSeekMath framework, replaces the traditional critic with group-based advantage estimation. 
Given a prompt $q$, the policy samples a group of $G$ outputs $\{o_1, \dots, o_G\}$. 
The advantage $\hat{A}_i$ of each output is computed by normalizing the rewards $\{r_1, \dots, r_G\}$ within the group:

\begin{equation}
\hat{A}_i = \frac{r_i - \mathrm{mean}(r_1, \dots, r_G)}{\mathrm{std}(r_1, \dots, r_G)}.
\end{equation}

The policy is optimized using a clipped surrogate objective with a KL-divergence regularization term to stabilize training.

\begin{table}[t]
\centering
\small
\setlength{\tabcolsep}{2pt} 
\begin{tabular}{cccccc}
\toprule
        & Examples & Databases & Tables & Domains & Rows \\
\midrule
BIRD & 12,751 & 92 & 7.3 & 37 & 549k \\
Spider & 10,181 & 200 & 5.1 & 138 & 2k \\
\bottomrule
\end{tabular}
\caption{Statistics of the BIRD and Spider benchmarks}
\label{table:dataset}
\end{table}

\subsubsection{Decoupled Clip and Dynamic Sampling Policy Optimization}

DAPO~\citep{dapo} extends GRPO with several techniques designed to improve exploration and sampling efficiency in long chain-of-thought (CoT) reasoning. 
The main components include:

\begin{enumerate}

\item \textbf{Clip-Higher.} 
An asymmetric clipping mechanism where the upper clipping bound $\epsilon_{\text{high}}$ (e.g., $0.28$) is larger than the lower bound $\epsilon_{\text{low}}$ (e.g., $0.20$), allowing more aggressive updates for high-reward samples.

\item \textbf{Dynamic Sampling.} 
Groups with zero reward variance are filtered out during training, ensuring that computation focuses on samples providing meaningful gradient signals.

\item \textbf{Token-Level Loss.} 
Gradients are aggregated at the token level to mitigate the bias toward shorter responses that may arise in long reasoning trajectories.

\end{enumerate}

\subsubsection{Group Sequence Policy Optimization}

Group Sequence Policy Optimization (GSPO)~\citep{gspo} addresses training instability in large Mixture-of-Experts (MoE) models. 
Instead of token-level optimization, GSPO operates at the sequence level. 
The importance ratio $s_i(\theta)$ is computed using length-normalized sequence likelihood:

\begin{equation}
s_i(\theta) =
\left(
\frac{\pi_\theta(y_i \mid x)}
{\pi_{\theta_{\text{old}}}(y_i \mid x)}
\right)^{\frac{1}{|y_i|}} .
\end{equation}

By shifting the optimization granularity from tokens to sequences, GSPO reduces the high variance introduced by token-level importance sampling in MoE architectures and improves training stability.

\begin{table}[t]
\centering
\small
\begin{tabular}{ll}
\toprule
\textbf{Parameter} & \textbf{Value} \\
\midrule
\multicolumn{2}{l}{\textit{Hardware Configuration}} \\
GPU (4B / 8B) & 8 $\times$ A100 (80GB) \\
GPU (30B-A3B MoE) & 32 $\times$ A100 (80GB) \\

\midrule
\multicolumn{2}{l}{\textit{Common Training Parameters}} \\
Max prompt length & 8192 \\
Max response length & 3072 \\
Max tool response length & 256 \\
Responses per prompt ($n_{\text{resp}}$) & 8 \\
Training prompt batch size & 128 \\
Max assistant turns & 10 \\
Learning rate & $1\times10^{-6}$ \\
Rollout temperature & 1.0 \\
Loss aggregation mode & token-mean \\
Rollout Type & vLLM~\citep{vllm} \\

\midrule
\multicolumn{2}{l}{\textit{GRPO}} \\
Clip ratio & 0.2 \\
KL regularization & Enabled \\

\midrule
\multicolumn{2}{l}{\textit{DAPO}} \\
Clip ratio (low, high) & (0.2, 0.28) \\
Overlong buffer length & 2048 \\
Dynamic sampling & Enabled \\
Token-level loss & Enabled \\

\midrule
\multicolumn{2}{l}{\textit{GSPO}} \\
Optimization granularity & Sequence-level \\
Length normalization & Enabled \\

\midrule
\multicolumn{2}{l}{\textit{GIGPO}} \\
Hierarchical advantage & Enabled \\
Step weight ($\omega$) & 1.0 \\

\midrule
\multicolumn{2}{l}{\textit{FineStep (Ours)}} \\
Step advantage weight ($\lambda$) & 0.5 \\
Reward smoothing ($\beta$) & 0.5 \\

\bottomrule
\end{tabular}
\caption{Training configuration and algorithm-specific hyperparameters.}
\label{tab:training_config}
\end{table}

\subsubsection{Group-in-Group Policy Optimization}

Group-in-Group Policy Optimization (GIGPO)~\citep{gigpo} introduces a hierarchical advantage structure for fine-grained credit assignment in multi-turn LLM agents. 
It groups trajectories that visit identical environment states (referred to as anchor states) and computes advantages using both global and local signals. 
The final advantage is defined as

\begin{equation}
A(a_t^{(i)}) = A_E(\tau_i) + \omega A_S(a_t^{(i)}),
\end{equation}

where $A_E$ denotes the episode-level relative advantage capturing task completion, and $A_S$ denotes the step-level relative advantage measuring local action effectiveness. 
However, both signals are still derived from outcome-level rewards and do not explicitly evaluate the utility of intermediate reasoning steps. 
As a result, compared with methods such as FineStep that directly model process-level rewards, the credit assignment in GIGPO remains relatively coarse-grained.

\subsection{More experimental Configuration.} \label{app:exp_set}

All RL algorithms are implemented using the \texttt{verl}~\citep{verl} framework under a unified training setup. 
For the 4B and 8B models, experiments are conducted on 8 NVIDIA A100 GPUs (80GB), while the MoE model (30B-A3B) is trained on 32 NVIDIA A100 GPUs (80GB). During inference, we fix the random seed for vLLM to ensure reproducibility under both greedy decoding and sampling settings. 
Implementation details can be found in Table~\ref{tab:training_config}.

\begin{table}[t]
\centering
\scriptsize 
\setlength{\tabcolsep}{1.2pt} 
\begin{tabular}{lccccc}
\toprule
\multirow{2}{*}{ Methods} & \multicolumn{2}{c}{ BIRD Dev }& \multicolumn{2}{c}{ Spider Dev } & \multicolumn{1}{c}{Spider Test}\\
    \cmidrule(r){2-3} \cmidrule(l){4-5} \cmidrule(l){6-6} 
    & EX  & VES & EX & TS &  EX \\
\midrule
\multicolumn{6}{l}{\textit{Closed-source methods}} \\
XiYan-SQL~\citep{xiyansql} & \textbf{73.3} & - & - & - & \textbf{89.6} \\
CHASE-SQL~\citep{pourreza2024chase} & 73.0 & \textbf{73.0} & - & - & 87.6 \\
MCTS-SQL + GPT-4~\citep{MCTS-SQL} & 69.4 & 66.2 & 88.7 & - & 86.6 \\
PAS-SQL + GPT-4o~\citep{PAS-SQL} & 64.7 & 65.0 & 87.9 & - & 86.8 \\
MCS-SQL + GPT-4~\citep{mcssql} & 63.4 & 64.8 & \textbf{89.5} & - & \textbf{89.6} \\
\midrule
\multicolumn{6}{l}{\textit{Training base methods}} \\
SFT CodeS-7B~\citep{DBLP:journals/pacmmod/codes} & 57.2 & 58.8 & 85.4 & 80.3 & - \\
OmniSQL-7B~\citep{OmniSQL} & 66.1 & - & 81.6 & - & 88.9 \\
Reasoning-SQL~\citep{reasoning-sql} & 64.0 & - & - & - & 78.7 \\
SQl-R1-7B~\citep{sql-r1} & 63.1 & - & 84.5 & - & 86.1 \\
Arcitc-Text2SQL-R1~\citep{Arctic-sql} & 64.8 & - & - & - & 87.1 \\
SQL-o1-8B~\citep{SQL-o1} & 63.4 & 64.7 & 87.4 & 79.6 & 85.4 \\
ReEx-SQL-7B~\citep{reex-sql} & 65.3 & \textbf{73.3} & 89.1 & 83.9 & 86.9 \\
MTIR-SQL + Qwen3-14B~\citep{mtir-sql} & 68.1 & -  &  86.7 & - &  87.2\\
\midrule
\multicolumn{6}{l}{\textit{Our method}} \\
{FineStep-4B} & \textbf{68.51}& {71.65} & \textbf{90.3}& \textbf{84.3}& \textbf{89.9} \\ 
{FineStep-8B} & {68.06} & {70.90} & {88.9} & {82.4} & \textbf{89.9}\\
\bottomrule
\end{tabular}
\caption{Performance comparison on BIRD and Spider benchmarks (\%).}
\label{table:exp1_indomain}
\end{table}

\begin{table}[t]
\centering
\scriptsize 
\setlength{\tabcolsep}{2.2pt} 
\begin{tabular}{lccccc}
\toprule
\multirow{2}{*}{Methods} & \multicolumn{2}{c}{Syn} & \multicolumn{2}{c}{Realistic} & \multicolumn{1}{c}{DK}\\
    \cmidrule(r){2-3} \cmidrule(l){4-5} \cmidrule(l){6-6}
    & EX  & TS  & EX & TS & EX \\
    \midrule
    \multicolumn{6}{l}{\textit{Closed-source methods}} \\
FastRAT\textsubscript{ext}+GPT-4~\citep{fastrat} & 74.4 & - & \textbf{80.9} & - & 72.3 \\
TA-SQL + GPT-4~\citep{TASQL} & - & - & 79.5 & - & \textbf{72.9} \\
CYCLESQL + GPT-4~\citep{CYCLESQL} & \textbf{76.0} & \textbf{66.3} & 70.6 & \textbf{56.9} & 68.5 \\
SQL-PaLM + PaLM2~\citep{SQLpalm}  & 74.6 & - & 77.6 & - & 66.5 \\
\midrule
\multicolumn{6}{l}{\textit{Training base methods}} \\
ROUTE + Llama3-8B~\citep{ROUTE} & 77.4 & 70.2 & 80.9 & 72.6 & 74.6 \\
SFT CodeS-15B~\citep{DBLP:journals/pacmmod/codes} & 77.0 & 69.4 & 83.1 & 75.6 & 70.7 \\
SENSE-13B~\citep{DBLP:SENSE} & 77.6 & 70.2 & 84.1 & 76.6 & 80.2 \\
OmniSQL-32B~\citep{OmniSQL} & 72.1 & - & 77.2 & - & 77.6 \\
Reasoning-SQL$^{*}$~\citep{reasoning-sql} & 69.3 & - & - & - & 73.3 \\
SQl-R1-7B$^{*}$~\citep{sql-r1} & 76.7 & - & 83.3 & - & 78.1 \\
SQL-o1 + Llama3-8B~\citep{SQL-o1} & 77.6 & 69.2 & 82.7 & 72.8 & 78.7 \\
ReEx-SQL-7B~\citep{reex-sql}& 78.7 & 72.0 & 85.5 & 79.9 & 80.2 \\
MTIR-SQL + Qwen3-14B~\citep{mtir-sql} & 81.0 & -  & 81.1 & - & 76.3 \\
\midrule
\multicolumn{6}{l}{\textit{Our method}} \\
FineStep-4B & \textbf{83.1} & \textbf{75.2} & \textbf{88.6} & \textbf{80.9} & \textbf{81.5}  \\
FineStep-8B & 81.4 & 71.6 & 86.0 & 77.9 & 80.0 \\
\bottomrule
\end{tabular}
\caption{Performance comparison on Spider variants robustness benchmarks (\%).}
\label{table:exp1_outdomain}
\end{table}

\subsection{Comparison with Existing Methods}

Tables~\ref{table:exp1_indomain} and \ref{table:exp1_outdomain} compare FineStep with a wide range of existing Text-to-SQL systems on both in-domain and out-of-domain benchmarks.

\textbf{In-domain performance.}
As shown in Table~\ref{table:exp1_indomain}, FineStep achieves competitive results against both closed-source and training-based approaches on the BIRD and Spider benchmarks. 
Notably, FineStep-4B achieves 90.3\% EX and 84.3\% TS on Spider Dev, outperforming many existing methods despite using a significantly smaller model. 
On Spider Test, both FineStep-4B and FineStep-8B reach 89.9\% EX, which is competitive with several GPT-4-based systems.
On the BIRD Dev, FineStep-4B also achieves strong performance, reaching 68.51\% EX and 71.65\% VES.

\textbf{Out-of-domain robustness.}
Table~\ref{table:exp1_outdomain} further evaluates model robustness on Spider variants, including Spider-Syn, Spider-Realistic, and Spider-DK.
FineStep consistently achieves state-of-the-art or competitive performance across these benchmarks.
For example, FineStep-4B obtains 81.4\% EX on Spider-Syn and 86.0\% EX on Spider-Realistic, outperforming many larger models.
FineStep-8B further achieves the best performance on Spider-DK with 81.9\% EX.

Overall, these results demonstrate that FineStep maintains strong performance not only on standard benchmarks but also under distribution shifts, highlighting its effectiveness in improving both reasoning accuracy and robustness.

\subsection{Theoretical Analysis of FineStep}\label{app:theory}

We provide a simple theoretical interpretation showing why trajectory-level outcome optimization may incorrectly reinforce useless intermediate step in successful trajectory, while FineStep can suppress them through step-level credit assignment.

\begin{proposition}[{Successful $\tau$ may contain useless $a_k$}]
Consider a multi-step reasoning trajectory
\begin{equation}
\tau = (s_0, a_0, s_1, \ldots, a_T, s_{T+1})
\end{equation}
with final outcome reward $R^{\mathrm{out}}(\tau)=1$. A step $a_k$ is \emph{useless} if removing it does not change the final answer or outcome:
\begin{equation}
R^{\mathrm{out}}(\tau) = R^{\mathrm{out}}(\tau \setminus a_k).
\end{equation}
Such steps naturally exist in tool-integrated reasoning. For example, an SQL tool call may be syntactically valid but irrelevant to the final query, or may repeat already obtained information. The trajectory can still be successful if later steps recover and produce the correct final SQL.
\end{proposition}

\begin{proposition}[{{GRPO reinforces $a_k$ in successful $\tau$}}]
In trajectory-level GRPO, all steps in the same trajectory share the same normalized outcome advantage:
\begin{equation}
A_{i,t}^{\mathrm{GRPO}} = \frac{R_i^{\mathrm{out}} - \mathrm{Mean}(\{R_j^{\mathrm{out}}\}_{j=1}^{G})}{\mathrm{Std}(\{R_j^{\mathrm{out}}\}_{j=1}^{G}) + \epsilon}.
\end{equation}
For a successful trajectory with above-average outcome reward, we have $A_{i,t}^{\mathrm{GRPO}} > 0, \forall t$.
The policy-gradient update for step $t$ is proportional to
\begin{equation}
A_{i,t}^{\mathrm{GRPO}} \nabla_\theta \log \pi_\theta(a_{i,t} \mid s_{i,t}).
\end{equation}
Therefore, if $a_{i,k}$ is a useless step inside a successful trajectory, it still receives a positive update $A_{i,k}^{\mathrm{GRPO}} > 0$.
Thus, GRPO cannot distinguish useful steps from useless ones within a successful trajectory, and will incorrectly increase the probability of both.
\end{proposition}

\begin{proposition}[{ FineStep can suppress useless $a_k$}]
FineStep assigns each step its own advantage:
\begin{equation}
A_{i,t}^{\mathrm{FineStep}} = \lambda \hat{A}^{\mathrm{out}}_{i,t} + (1-\lambda)\hat{A}^{\mathrm{proc}}_{i,t}.
\end{equation}
For a useless step $a_{i,k}$, its process value is low because it provides little or no information gain:
\begin{equation}
V^{\mathrm{proc}}_{i,k} < \mathrm{Mean}(\mathcal{V}^{\mathrm{proc}}),
\end{equation}
which leads to $\hat{A}^{\mathrm{proc}}_{i,k} < 0$.
FineStep suppresses this step when
\begin{equation}
\lambda \hat{A}^{\mathrm{out}}_{i,k} + (1-\lambda)\hat{A}^{\mathrm{proc}}_{i,k} < 0,
\end{equation}
equivalently,
\begin{equation}
|\hat{A}^{\mathrm{proc}}_{i,k}| > \frac{\lambda}{1-\lambda} \hat{A}^{\mathrm{out}}_{i,k}.
\end{equation}
Under this condition, the policy-gradient update for the useless step becomes negative:
\begin{equation}
A_{i,k}^{\mathrm{FineStep}} \nabla_\theta \log \pi_\theta(a_{i,k} \mid s_{i,k}), \quad A_{i,k}^{\mathrm{FineStep}} < 0,
\end{equation}
which decreases the probability of the useless action.
\end{proposition}

This analysis shows that trajectory-level GRPO gives identical credit to all steps in a successful trajectory, including redundant or uninformative ones. In contrast, FineStep can assign lower or even negative advantages to useless steps by incorporating step-level process signals. Therefore, FineStep provides a more discriminative credit assignment mechanism for long-horizon tool-integrated reasoning.

\subsection{Computational Overhead Analysis}\label{app:overhead}

We demonstrate the computational overhead in both training and inference scenarios. Our results show that the gains of FineStep do not come from longer reasoning, more tool calls, or higher test-time computation, but from more efficient step-level credit assignment.

\subsubsection{Training Overhead Analysis}

From the training dynamics as shown in Figure~\ref{fig:exp22_ablation_of_components}, different RL methods exhibit distinct overhead:
\begin{itemize}[leftmargin=1.5em, itemsep=2pt]
    \item \textbf{GSPO} tends to produce more interaction turns, leading to substantially more tool calls.
    \item \textbf{GRPO} and \textbf{DAPO} rely more on longer reasoning text, resulting in larger response lengths.
    \item \textbf{GIGPO} and \textbf{FineStep} obtain external feedback through more stable tool usage and shorter reasoning traces.
\end{itemize}
By assigning credit at the step level, FineStep suppresses redundant reasoning and invalid tool interactions. As a result, it improves accuracy without noticeably increasing rollout overhead.

\subsubsection{Inference Overhead Analysis}

We further measure inference overhead on the BIRD Dev set using Qwen3-4B-Instruct as the base model. Table~\ref{tab:overhead_greedy} reports results under greedy decoding, and Table~\ref{tab:overhead_sc} reports results under self-consistency with 8 samples.  In both tables, Tool Calls, Avg. Time, and Resp. Len. denote the average number of tool calls, average inference time, and average response length per sample, respectively.

\begin{table}[t]
\centering
\small
\setlength{\tabcolsep}{3pt} 
\begin{tabular}{lcccc}
\toprule
Method & EX(\%) & Tool Calls & Avg. Time (s) & Resp. Len. \\
\midrule
GRPO & 65.78 & 1.09 & 0.231 & 841.7 \\
DAPO & 66.75 & 1.10 & 0.276 & 1086.2 \\
GSPO & 67.01 & 3.66 & 0.421 & 732.4 \\
GIGPO & 66.82 & 1.23 & 0.252 & 552.2 \\
FineStep & \textbf{67.01} & 1.22 & 0.250 & 541.4 \\
\bottomrule
\end{tabular}
\caption{Inference overhead under greedy decoding.}
\label{tab:overhead_greedy}
\end{table}

\begin{table}[t]
\centering
\setlength{\tabcolsep}{3pt} 
\small
\begin{tabular}{lcccc}
\toprule
Method & EX(\%) & Tool Calls & Avg. Time (s) & Resp. Len. \\
\midrule
GRPO & 66.23 & 8.71 & 1.064 & 6728 \\
DAPO & 67.67 & 8.83 & 1.295 & 8688 \\
GSPO & 68.06 & 29.27 & 2.981 & 5856 \\
GIGPO & 67.99 & 9.84 & 1.174 & 4416 \\
FineStep & \textbf{68.51} & 9.73 & 1.162 & 4328 \\
\bottomrule
\end{tabular}
\caption{Inference overhead under self-consistency with 8 samples.}
\label{tab:overhead_sc}
\end{table}

FineStep achieves the highest EX under self-consistency, while maintaining tool calls, inference time, and response length comparable to GIGPO and substantially lower than GSPO.
Compared with GSPO, FineStep improves EX from 68.06 to 68.51, while reducing tool calls from 29.27 to 9.73 and average inference time from 2.981s to 1.162s.
These results indicate that FineStep's improvements do not rely on increased test-time computation, but instead come from more efficient and reliable tool-integrated reasoning.

\subsection{Generalizability on Non-SQL Reasoning}\label{app:generalizability}

To evaluate the generalizability of FineStep beyond Text-to-SQL, we conduct a preliminary experiment on AIME24, a mathematical reasoning benchmark, using Qwen2.5-7B-Instruct as the base model with a Python execution tool for interactive reasoning. Results are shown in Table~\ref{tab:aime24}.

\begin{table}[h]
\centering
\small
\begin{tabular}{lc}
\toprule
Method & Pass@1 Accuracy \\
\midrule
No-Train & 16.7 \\
GRPO & 33.3 \\
FineStep & \textbf{34.6} \\
\bottomrule
\end{tabular}
\caption{Results on AIME24 (Qwen2.5-7B-Instruct).}
\label{tab:aime24}
\end{table}

Since AIME24 has no SQL executor or intermediate execution results, we adapt FineStep with the following modifications:
\begin{enumerate}[leftmargin=1.5em, itemsep=2pt]
    \item \textbf{No SQL-specific soft reward.} The recall-based reward used in Text-to-SQL is not applicable, as there is no intermediate execution feedback.
    \item \textbf{Only hard-constraint process reward.} We use simple process constraints to transfer the step-level credit assignment idea without introducing task-specific reward shaping.
    \item \textbf{No process smoothing.} The process signal in AIME24 is weaker than SQL execution feedback; smoothing may propagate noisy signals.
    \item \textbf{Larger outcome weight ($\lambda=0.8$).} Since final-answer correctness is more reliable than the weak process signal, we let the outcome reward dominate while keeping the process reward as an auxiliary constraint.
\end{enumerate}

These preliminary results suggest that FineStep's step-level credit assignment can provide gains beyond Text-to-SQL, even with minimal task-specific adaptation. Broader general-agent evaluation remains future work.

\subsection{Reward Robustness and Hyperparameter Sensitivity}\label{app:hyperparam}

FineStep's reward design relies only on general SQL executor feedback, without benchmark-specific rules or annotations. Therefore, its gains mainly come from step-level credit assignment rather than benchmark-specific reward tuning.

To further verify this, we conduct cross-dataset hyperparameter sensitivity analysis. Using Qwen3-4B-Instruct, we fix $\lambda=0.5$ and vary $\beta$, and then fix $\beta=0.5$ and vary $\lambda$. We report results on BIRD Dev, Spider Dev, and Spider-Realistic.

\subsubsection{Sensitivity to $\beta$ ($\lambda=0.5$)}

\begin{table}[h]
\centering
\small
\setlength{\tabcolsep}{3pt}
\begin{tabular}{lccccccc}
\toprule
$\beta$ & 0 & 0.2 & 0.4 & 0.5 & 0.6 & 1 \\
\midrule
BIRD Dev & 65.84 & 66.82 & 67.86 & \textbf{68.51} & 68.19 & 66.36 \\
Spider Dev & 87.3 & 88.4 & 89.9 & \textbf{90.3} & 89.7 & 87.2 \\
Spider-Realistic & 86.9 & 87.6 & 87.4 & \textbf{88.6} & 88.2 & 86.0 \\
\bottomrule
\end{tabular}
\caption{EX (\%) with varying $\beta$ (fixed $\lambda=0.5$).}
\label{tab:sensitivity_beta}
\end{table}

\subsubsection{Sensitivity to $\lambda$ ($\beta=0.5$)}

\begin{table}[h]
\centering
\small
\setlength{\tabcolsep}{3pt}
\begin{tabular}{lccccccc}
\toprule
$\lambda$ & 0 & 0.2 & 0.4 & 0.5 & 0.6 & 1 \\
\midrule
BIRD Dev & 67.41 & 67.21 & 67.73 & \textbf{68.51} & 67.60  & 65.97 \\
Spider Dev & 88.2 & 88.3 & 89.5 & \textbf{90.3} & 88.9  & 87.6 \\
Spider-Realistic & 87.4 & 87.6 & 88.1 & \textbf{88.6} & 87.8  & 85.4 \\
\bottomrule
\end{tabular}
\caption{EX (\%) with varying $\lambda$ (fixed $\beta=0.5$).}
\label{tab:sensitivity_lambda}
\end{table}

The results show that $\beta=0.5$ and $\lambda=0.5$ achieve the best or near-best performance not only on BIRD Dev, but also on Spider Dev and Spider-Realistic. This demonstrates FineStep's robustness to reward design choices and its consistent cross-dataset generalization.

\subsection{Prompt Design}

FineStep Prompt (Figure \ref{prompt:finestepprompt}) Following the paradigm of MTIR-SQL~\citep{mtir-sql}, we designed a multi-turn tool-integrated reasoning prompt to elicit structured and verifiable logic. This design facilitates an iterative "hypothesize-verify-refine" process by providing distinct modules for <reasoning>, tool-based validation, and final answer encapsulation. Such a mechanism effectively reduces hallucinations in complex SQL generation by forcing the model to validate its internal reasoning against the database engine before committing to a final output.



\subsection{Case Study}
\begin{figure*}[htbp]
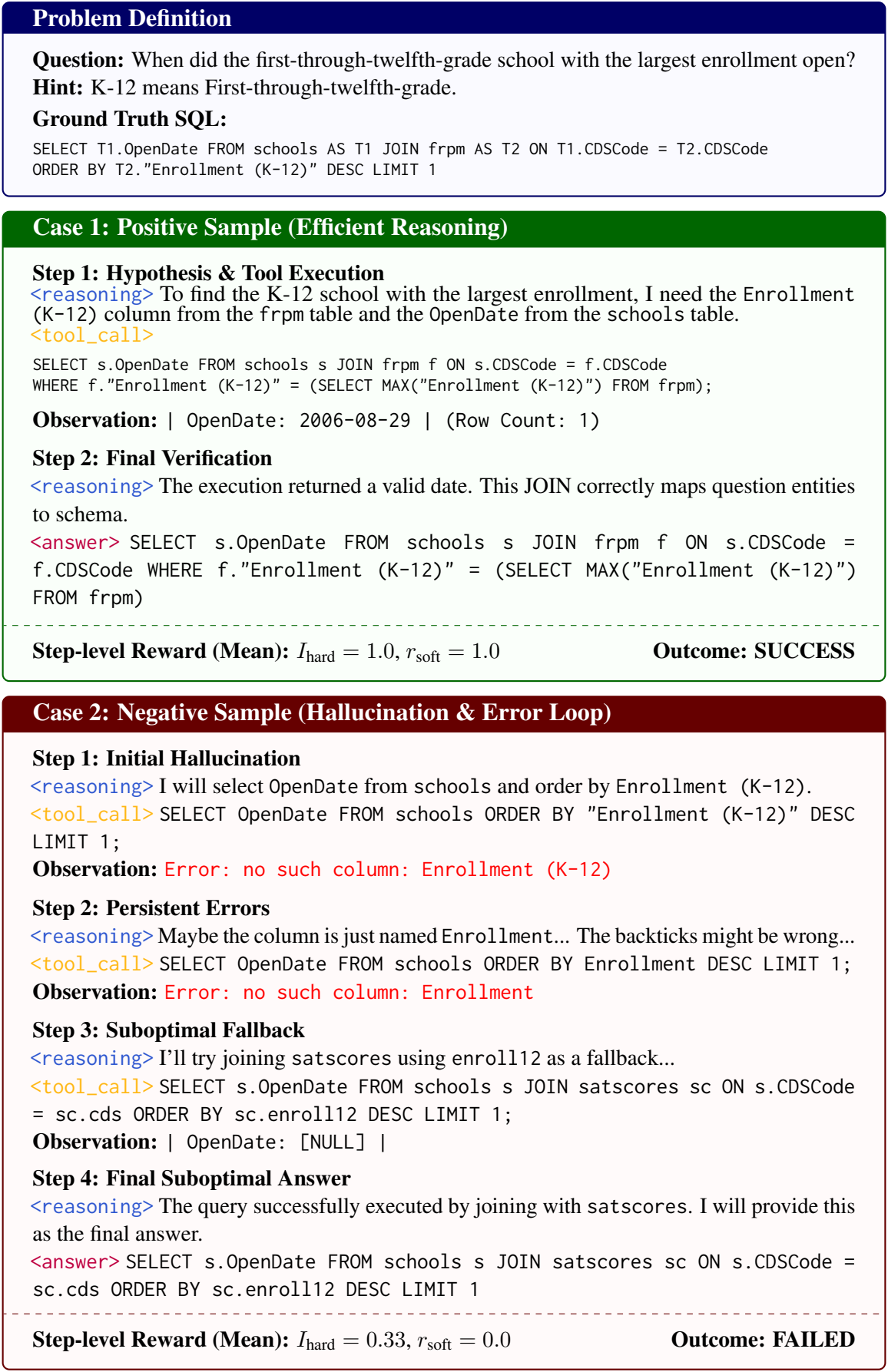

\centering
\begin{tcolorbox}[
    width=0.95\textwidth,
    title={Problem Definition}, 
    colback=blue!2, 
    colframe=blue!40!black, 
    fonttitle=\bfseries\large,
    enhanced,
    boxrule=0.8pt
]
\textbf{Question:} When did the first-through-twelfth-grade school with the largest enrollment open? \\
\textbf{Hint:} K-12 means First-through-twelfth-grade.

\vspace{0.5em}
\textbf{Ground Truth SQL:}
\begin{small}
\begin{verbatim}
SELECT T1.OpenDate FROM schools AS T1 JOIN frpm AS T2 ON T1.CDSCode = T2.CDSCode 
ORDER BY T2."Enrollment (K-12)" DESC LIMIT 1
\end{verbatim}
\end{small}
\end{tcolorbox}

\begin{tcolorbox}[
    width=0.95\textwidth,
    title={Case 1: Positive Sample (Efficient Reasoning)}, 
    colback=green!2, 
    colframe=green!40!black, 
    fonttitle=\bfseries\large,
    enhanced,
    boxrule=0.8pt
]
\textbf{Step 1: Hypothesis \& Tool Execution} \\
\textcolor[HTML]{2e57d0}{\texttt{<reasoning>}} To find the K-12 school with the largest enrollment, I need the \texttt{Enrollment (K-12)} column from the \texttt{frpm} table and the \texttt{OpenDate} from the \texttt{schools} table. \\
\textcolor[HTML]{ffb713}{\texttt{<tool\_call>}} 
\begin{small}
\begin{verbatim}
SELECT s.OpenDate FROM schools s JOIN frpm f ON s.CDSCode = f.CDSCode 
WHERE f."Enrollment (K-12)" = (SELECT MAX("Enrollment (K-12)") FROM frpm);
\end{verbatim}
\end{small}
\textbf{Observation:} \texttt{| OpenDate: 2006-08-29 | (Row Count: 1)}

\vspace{0.5em}
\textbf{Step 2: Final Verification} \\
\textcolor[HTML]{2e57d0}{\texttt{<reasoning>}} The execution returned a valid date. This JOIN correctly maps question entities to schema. \\
\textcolor[HTML]{bf0040}{\texttt{<answer>}} \texttt{SELECT s.OpenDate FROM schools s JOIN frpm f ON s.CDSCode = f.CDSCode 
WHERE f."Enrollment (K-12)" = (SELECT MAX("Enrollment (K-12)") FROM frpm)}
\tcbline
\textbf{Step-level Reward (Mean):} $I_{\text{hard}}=1.0$, $r_{\text{soft}}=1.0$ \hfill \textbf{Outcome: SUCCESS}
\end{tcolorbox}

\begin{tcolorbox}[
    width=0.95\textwidth,
    title={Case 2: Negative Sample (Hallucination \& Error Loop)}, 
    colback=red!2, 
    colframe=red!40!black, 
    fonttitle=\bfseries\large,
    enhanced,
    boxrule=0.8pt
]
\textbf{Step 1: Initial Hallucination} \\
\textcolor[HTML]{2e57d0}{\texttt{<reasoning>}} I will select \texttt{OpenDate} from \texttt{schools} and order by \texttt{Enrollment (K-12)}. \\
\textcolor[HTML]{ffb713}{\texttt{<tool\_call>}} \texttt{SELECT OpenDate FROM schools ORDER BY "Enrollment (K-12)" DESC LIMIT 1;} \\
\textbf{Observation:} \textcolor{red}{\texttt{Error: no such column: Enrollment (K-12)}}

\vspace{0.5em}
\textbf{Step 2: Persistent Errors} \\
\textcolor[HTML]{2e57d0}{\texttt{<reasoning>}} Maybe the column is just named \texttt{Enrollment}... The backticks might be wrong... \\
\textcolor[HTML]{ffb713}{\texttt{<tool\_call>}} \texttt{SELECT OpenDate FROM schools ORDER BY Enrollment DESC LIMIT 1;} \\
\textbf{Observation:} \textcolor{red}{\texttt{Error: no such column: Enrollment}}

\vspace{0.5em}
\textbf{Step 3: Suboptimal Fallback} \\
\textcolor[HTML]{2e57d0}{\texttt{<reasoning>}} I'll try joining \texttt{satscores} using \texttt{enroll12} as a fallback... \\
\textcolor[HTML]{ffb713}{\texttt{<tool\_call>}} \texttt{SELECT s.OpenDate FROM schools s JOIN satscores sc ON s.CDSCode = sc.cds ORDER BY sc.enroll12 DESC LIMIT 1;} \\
\textbf{Observation:} \texttt{| OpenDate: [NULL] |} 

\vspace{0.4em}
\textbf{Step 4: Final Suboptimal Answer} \\
\textcolor[HTML]{2e57d0}{\texttt{<reasoning>}} The query successfully executed by joining with \texttt{satscores}. I will provide this as the final answer. \\
\textcolor[HTML]{bf0040}{\texttt{<answer>}} \texttt{SELECT s.OpenDate FROM schools s JOIN satscores sc ON s.CDSCode = sc.cds ORDER BY sc.enroll12 DESC LIMIT 1}

\tcbline
\textbf{Step-level Reward (Mean):} $I_{\text{hard}}=0.33$, $r_{\text{soft}}=0.0$ \hfill \textbf{Outcome: FAILED}
\end{tcolorbox}

\caption{Comparison of reasoning paths. The top block shows the task definition and Ground Truth; the middle and bottom blocks contrast a successful FineStep trajectory with a failed hallucination-prone trajectory.}
\label{fig:case_study_three_blocks}
\end{figure*}

Figure~\ref{fig:case_study_three_blocks} presents a representative example from the BIRD development set (question\_id = 29) to illustrate the behavioral differences between FineStep and GRPO. 
The middle block shows a successful reasoning trajectory produced by FineStep. 
With explicit step-level supervision, the model correctly identifies the relevant tables (\texttt{schools} and \texttt{frpm}) and generates a valid SQL query in a single interaction step, leading to the correct execution result.

In contrast, the bottom block shows a trajectory generated by a GRPO-trained model. 
Without explicit process-level guidance, the model initially hallucinates a nonexistent column and repeatedly attempts invalid queries, eventually falling back to an unrelated table (\texttt{satscores}). 
Although the final query executes successfully, it does not correspond to the correct reasoning path and produces an incorrect result.

This example highlights the advantage of FineStep in guiding models toward valid and informative intermediate steps, thereby avoiding hallucination loops and improving reasoning reliability.

\begin{figure*}[htbp]
\begin{tcolorbox}[title={Multi-Turn Tool-Integrated Reasoning Prompt},colback=yellow!1.5]
\textbf{Task Overview:} \\
You are a helpful SQL expert assistant. You should first think about how to write the SQL query by analyzing the question, database schema, and external knowledge, you can use the provided SQL execution tool to verify and refine your reasoning hypotheses. Finally, you provide the final SQL query in \texttt{<answer> </answer>}.\\
\rule{\linewidth}{0.4pt}

\textbf{Database Information:}
\begin{itemize}[leftmargin=1em, itemsep=0pt]
    \item \textbf{Database Engine:} \{dialect\}
    \item \textbf{Database Schema:} \{db\_schemas\}
    \item \textbf{Matched Contents:} \{matched\_contents\}
\end{itemize}

\textbf{Requirements:}
\begin{enumerate}[leftmargin=1.5em]
  \item \textbf{Precision:} Make sure you only output the information that is asked in the question. If the question asks for a specific column, make sure to only include that column in the SELECT clause, nothing more.
  \item \textbf{Completeness:} The generated query should return all of the information asked in the question without any missing or extra information.
  \item \textbf{Correctness:} Before generating the final SQL query, please think through the steps of how to write the query. Validate your reasoning through tool testing.
\end{enumerate}
\rule{\linewidth}{0.4pt}

\textbf{Output Format [Important]:} \\
Respond strictly in one of the following two modes. Do not mix the structures:

\vspace{0.5em}
\textbf{Option A (Tool call for validation or exploration):} \\
\textcolor[HTML]{2e57d0}{{<reasoning>}} Concise reasoning here. \textcolor[HTML]{2e57d0}{{</reasoning>}} \\
\texttt{[Tool call code/statement here]}

\vspace{0.5em}
\textbf{Option B (Final answer):} \\
\textcolor[HTML]{2e57d0}{{<reasoning>}} Concise reasoning here. \textcolor[HTML]{2e57d0}{{</reasoning>}} \\
\textcolor[HTML]{bf0040}{{<answer>}} YOUR FINAL SQL QUERY HERE \textcolor[HTML]{bf0040}{{</answer>}} \\
\rule{\linewidth}{0.4pt}

\textbf{[Question]} \{question\} \\
\textbf{[Hint]} \{hint\}
\end{tcolorbox}
\caption{The proposed multi-turn reasoning prompt for FineStep.}
\label{prompt:finestepprompt}
\end{figure*}

\end{document}